\documentclass[conference]{IEEEtran}% For LaTeX2e
\usepackage{cite}
\usepackage{amsmath,amssymb,amsfonts}
\usepackage{graphicx}
\usepackage{textcomp}
\usepackage[table]{xcolor}
\usepackage{booktabs}
\usepackage{multirow}
\usepackage{algorithm}
\usepackage{algpseudocode}
\usepackage{subcaption}
\usepackage[most]{tcolorbox}
\usepackage{xspace}
\usepackage{url}
\usepackage[bookmarks=false]{hyperref}

\usepackage{amsmath,amsfonts,bm}

\def\eqref#1{equation~\ref{#1}}
\def\1{\bm{1}}

\DeclareMathAlphabet{\mathsfit}{\encodingdefault}{\sfdefault}{m}{sl}
\SetMathAlphabet{\mathsfit}{bold}{\encodingdefault}{\sfdefault}{bx}{n}

\newtcolorbox{promptbox}[1][]{
    enhanced,
    breakable,
    colback=gray!4,
    colframe=gray!35,
    boxrule=0.6pt,
    arc=2pt,
    left=6pt,
    right=6pt,
    top=5pt,
    bottom=5pt,
    fontupper=\small,
    title=#1,
    fonttitle=\small\bfseries,
    coltitle=black,
    colbacktitle=gray!12,
    attach boxed title to top left={
        xshift=5pt,
        yshift=-2mm
    },
    boxed title style={
        boxrule=0pt,
        arc=1pt
    }
}
\makeatletter
\renewcommand{\paragraph}[1]{%
  \par\addvspace{2pt}%
  \noindent\textbf{#1}\hspace{0.5em}\ignorespaces}
\makeatother

\hypersetup{
    colorlinks=true,
    citecolor=blue,
    linkcolor=blue,
    urlcolor=blue
}

\newcommand{\name}{CARE\xspace}
\newcommand{\capthead}[1]{\textbf{#1}}

\title{Know When to Escalate: Critic-Guided Adaptive Execution for Cloud--Local Embodied Agents}

\author{
\IEEEauthorblockN{
Wei Zhang\IEEEauthorrefmark{1},
Lirong Xiang\IEEEauthorrefmark{2},
Charles Fleming\IEEEauthorrefmark{3}
}
\IEEEauthorblockA{
\IEEEauthorrefmark{1}University of Illinois Urbana-Champaign
\quad
\IEEEauthorrefmark{2}Cornell University
\quad
\IEEEauthorrefmark{3}Cisco
}
}

\begin{document}

\maketitle

\begin{abstract}
Recent advances in foundation-model agents have enabled increasingly capable embodied systems for long-horizon planning and interaction. However, embodied agents require fast, closed-loop decisions as they continuously perceive and act on a changing environment. Deploying strong cloud agents leaves the platform idle for seconds between consecutive actions, while lightweight local models respond quickly but are more vulnerable to errors. Existing hierarchical architectures typically assign models fixed roles, whereas conventional model routers make stateless per-request decisions.
We present \name, a critic-guided adaptive execution framework that dynamically determines \emph{when to escalate} control from a local executor to a stronger cloud model. A lightweight critic model monitors execution reliability and triggers escalation when local execution becomes unreliable; once escalated, control persists until the agent recovers, avoiding step-wise switching. To support such control transfer, a shared handoff memory maintains execution context and task-level progress across executors, enabling the model to intervene without restarting from scratch.
Across diverse embodied benchmarks, \name improves task success by $6.3$--$18.3$ percentage points over the strongest evaluated collaboration baseline. \name further approaches or exceeds Cloud-only success while reducing median time-between-action latency by $28.8$--$82.6\%$.
%Code is available at \url{https://anonymous.4open.science/r/anonymous-artifact-E02D/}.
\end{abstract}


\section{Introduction}
Executing long-horizon robotic tasks requires embodied agents to repeatedly interpret observations, select grounded actions, and update their decisions based on action outcomes. Recent advances in generative foundation models have strengthened these capabilities~\cite{OpenVLA, RL-GPT, EmbodiedGPT, RoboMP2}, enabling such models to coordinate perception, decision making, and embodied skills as general-purpose reasoning modules~\cite{SayCan, Code-as-policies, InnerMonologue, VLAs-as-Tools}. Because model inference lies directly within the perception--action loop, high inference latency can leave the agent acting on stale observations, potentially causing failure in time-sensitive interactions. To balance strong reasoning with responsive closed-loop interaction, recent hierarchical and slow--fast architectures pair powerful foundation models with lightweight local executors or specialized controllers~\cite{PARTNR, LARM}.

Despite these advances, long-horizon embodied execution remains challenging because each action affects what the agent observes and must do next. Consider an agent midway through a rearrangement task. An earlier action has displaced the target object, causing the next grasp to fail even though the resulting observation appears similar to the previous one. If the agent overlooks this feedback, it may repeat the same action, lose track of task progress, and compound earlier errors~\cite{PARTNR, EmbodiedAgentInterface}. Execution difficulty also changes along the trajectory. A lightweight executor may handle routine steps well but struggle after ambiguous observations, failed actions, or deviations from the expected state. The appropriate executor may therefore change as interaction unfolds.

\begin{figure}
    \centering
    \includegraphics[width=\linewidth]
    {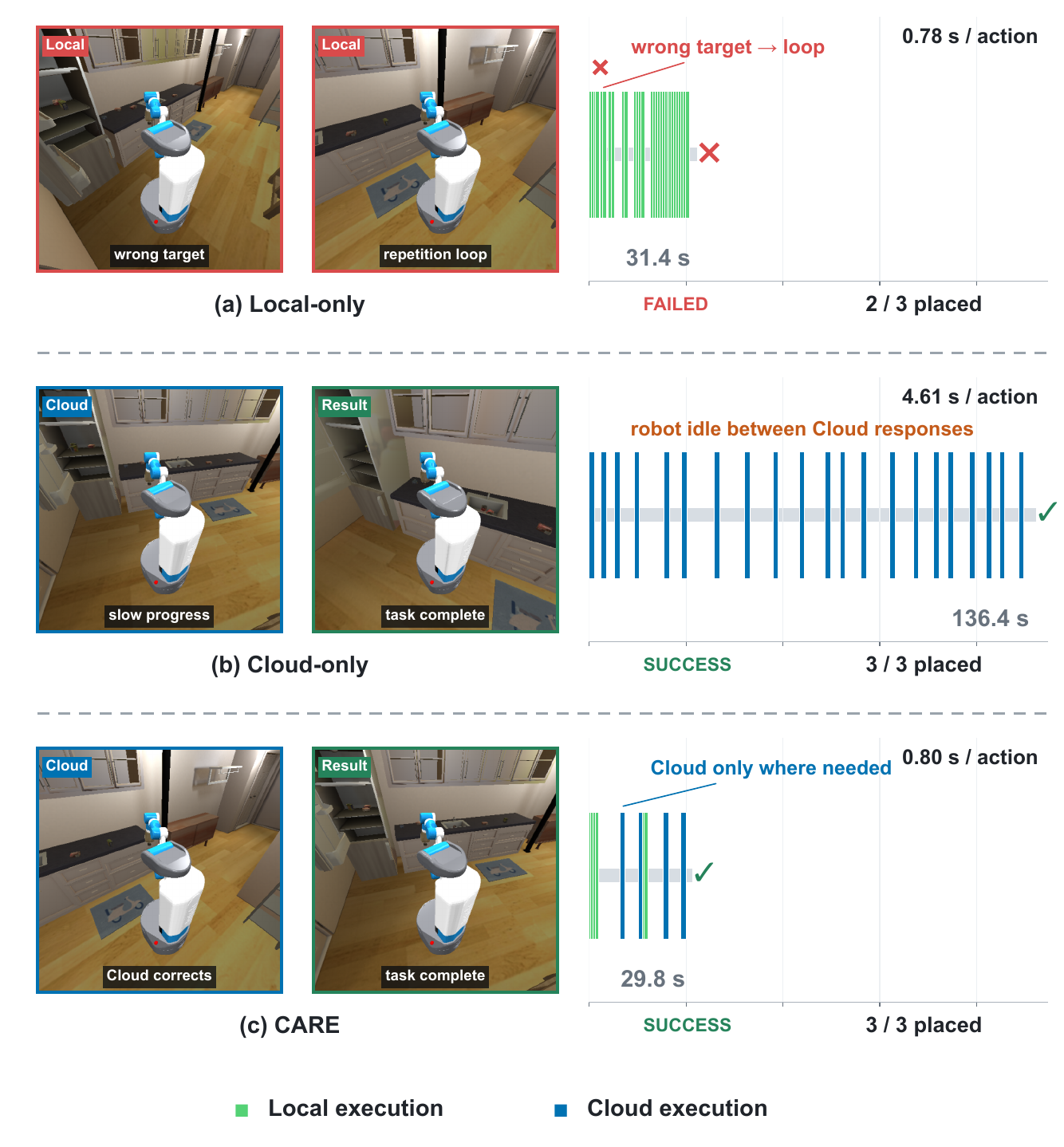}
    \caption{\capthead{No fixed assignment of executors fits a whole task.} The same rearrangement instance executed by (a) the local model only, (b) the cloud model only, and (c) \name. By adapting the executor as interaction unfolds, \name completes the task while retaining responsive execution.}
    \label{fig:introduction}
\end{figure}

This challenge is particularly pronounced in cloud--local embodied systems, where execution must continuously balance model capability against interaction latency. As shown in Figure~\ref{fig:introduction}, local models provide responsive interaction but may struggle in difficult states, whereas cloud models offer stronger reasoning at the cost of seconds per decision~\cite{RL-GPT, LARM, OpenVLA}. Existing hierarchical systems typically assign these models fixed planner--executor roles, implicitly assuming that the designated executor remains appropriate throughout the task. Replanning methods can revise plans or actions in response to feedback, but generally do so without reconsidering which model should execute the subsequent trajectory~\cite{InnerMonologue, AdaPlanner, ReflectivePlanning}. Per-query model routing likewise selects an executor independently for each request, without explicitly modeling how executor suitability evolves within an embodied trajectory~\cite{Routellm}.

We present \textsc{\name}, a critic-guided adaptive execution framework for cloud--local embodied agents. Rather than routing individual actions independently, \textsc{\name} uses a critic model to assess whether local execution should continue under the current interaction state and selectively escalates execution to the cloud when local execution becomes unreliable. Importantly, escalation is treated as a stateful control transfer rather than a one-step model selection: cloud intervention persists through the current recovery, avoiding premature hand-backs. Once the critic supports local execution again, control returns to the local executor to recover low-latency interaction. Meanwhile, a shared handoff memory preserves observations, completed progress, and failure context across model handoffs, allowing both executors to maintain a consistent execution state. By combining critic-guided control transfer with stateful handoff memory, \textsc{\name} avoids both persistent cloud invocation and myopic per-action routing while adapting execution authority to the evolving difficulty of the task.

Our main contributions are summarized as follows:

\begin{itemize}

\item We introduce a new formulation of cloud--local embodied collaboration as a stateful adaptive execution problem and propose a novel critic-guided execution framework that dynamically determines when execution should be escalated and when local control can be safely resumed.

\item We develop \name, a novel stateful execution mechanism that combines persistent cloud intervention with shared execution memory, enabling reliable control transfer while preserving execution continuity across model handoffs.

\item We conduct a comprehensive evaluation of \name across three long-horizon embodied benchmarks spanning diverse observation modalities and execution interfaces. \name improves task success by $6.3$--$18.3$ percentage points over the strongest evaluated collaboration baseline while reducing median time-between-action latency by $28.8$--$82.6\%$.

\end{itemize}

\section{Related Work}

\subsection{Long-horizon embodied agents}
Generative foundation models are increasingly used as high-level planners for embodied agents. Early work studies text-based interactive environments~\cite{SayCan, ALFWorld} using explicit world knowledge~\cite{WKM}, reflective reasoning~\cite{ReflectivePlanning, ReAct}, and plan construction~\cite{Progprompt}; with vision-language models, planning has extended to visually grounded interaction through scene graph representations~\cite{SceneGraphPlanning}, memory-augmented visual reasoning~\cite{Memoryvla}, and multimodal plans that guide downstream control policies~\cite{VLAs-as-Tools}. Other work trains open-source planners by reinforcement learning~\cite{bo2025reinforced} or supervised finetuning~\cite{Embodied-reasoner}. Another parallel line of work makes interaction latency a first-class design constraint by organizing execution hierarchically, pairing a computationally intensive foundation model with faster specialized models or policies. This hierarchy is instantiated through program or skill generation~\cite{Code-as-policies}, intermediate representation prediction~\cite{Pivot-r}, slow--fast reasoning and control~\cite{RL-GPT}, and semantic guidance to downstream policies~\cite{Rt-2}, while the execution models themselves have grown more capable through large-scale vision-language-action pretraining~\cite{OpenVLA, pi_0.5}. Both lines strengthen what each model contributes, but they fix the assignment in advance: planning improves the instructions an executor receives, and hierarchy fixes which model executes them. We instead study the runtime question left open by both---which model should hold execution authority as a plan is carried out, and when that authority should change.

\subsection{Embodied execution monitoring}
Outside embodied settings, cascades invoke a strong model only when a weak model's output fails a reliability check, using signals such as answer consistency~\cite{FrugalGPT, LLMCascade} or few-shot self-verification~\cite{AutoMix}, while routers predict the target model before generation~\cite{Routellm}. These methods decide over a single self-contained input: the escalated model answers once, and the decision does not affect what is asked next. Embodied execution violates both properties---control may later return to the local executor, and an incorrect action changes the state at which later decisions are made---so escalation becomes a stateful control problem rather than a routing problem. Within embodied systems, failure-recovery methods use execution feedback to diagnose failures and generate corrective plans~\cite{Liu2023, DoReMi}. Their primary output is an explanation or a revised plan. Our critic instead controls a different decision variable: which of two executors holds execution authority, and when that authority should change.

\section{Method}
\label{sec:method}

\begin{figure*}[t]
    \centering
    \includegraphics[width=\linewidth]
    {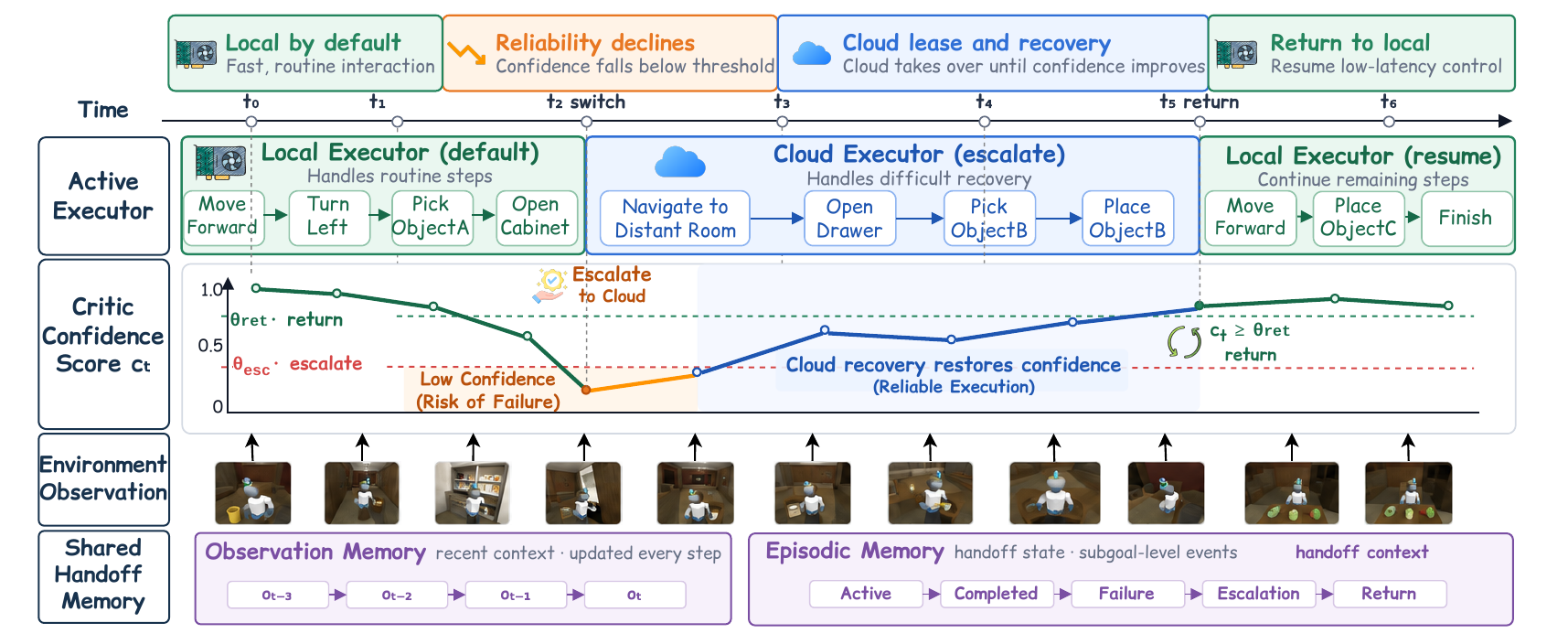}
    \caption{\capthead{Critic-guided control transfer over an execution trajectory.} Declining local-continuation confidence triggers persistent cloud takeover, while a higher return threshold restores local control after recovery.}
    \label{fig:timeline}
\end{figure*}

Figure~\ref{fig:timeline} illustrates \name along a single execution trajectory. The agent begins under local control. When a pick action repeatedly fails and progress stalls, \name changes which executor proposes the next action while leaving the perception--action loop unchanged. A critic assesses whether local execution should continue based on the current observation, local proposal, and execution history. Control transfers to the cloud when needed and remains there until the critic supports returning to local execution. Throughout these switches, shared handoff memory preserves task progress, allowing the incoming executor to continue from the current state rather than restart. We first formalize stateful cloud--local execution (\S\ref{sec:formulation}), then describe the critic-guided transfer mechanism (\S\ref{sec:switch}) and shared handoff memory (\S\ref{sec:memory}).

\subsection{Problem Formulation}
\label{sec:formulation}
We consider cloud--local collaboration at the embodied decision layer, where models select grounded actions while low-level motor control is handled by the underlying platform. The agent separates one-time task planning from closed-loop execution. Given a task instruction $G$ and initial observation $o_0$, a planner $\pi_P$ first produces a high-level plan $P = \pi_P(G,o_0)$, which provides task-level guidance throughout the episode. 
%Our focus is not on improving the planner itself, but on determining how execution control should adapt between models with different capability and latency characteristics.
The resulting plan is carried out by two executors: a lightweight local model $\pi_L$ and a high-capability cloud model $\pi_C$. Both operate at the same level of abstraction and output navigation or manipulation actions, but exhibit a capability--latency asymmetry: the cloud executor generally provides stronger reasoning at substantially higher inference latency. At interaction step $t$, both executors have access to a shared execution context:
\begin{equation}
    x_t = \bigl(G, P, o_t, M_t\bigr),
    \label{eq:execution_context}
\end{equation}
where $G$ denotes the task instruction, $o_t$ the current observation, and $M_t$ the shared execution memory accumulated along the trajectory.

Let $z_t\in\{L,C\}$ denote the executor in control at step $t$. The executed action is then denoted as $a_t = \pi_{z_t}(x_t)$. Executing $a_t$ yields the next observation $o_{t+1}$ together with an execution outcome $e_t$, which captures environment feedback from the interaction. The control state $z_t$ evolves over time according to the critic-guided transfer mechanism introduced in \S\ref{sec:switch}.

Our objective is to approach Cloud-only task success while reducing closed-loop response time, particularly the delay between consecutive embodied actions. Because each executed action changes the subsequent observation and task state, executor selection is coupled across the trajectory rather than made independently for each request.

\subsection{Critic-Guided Control Transfer}
\label{sec:switch}

\paragraph{Assessing Local Execution.}
The central challenge is to determine when the local executor should retain control and when execution should be escalated to the cloud. This judgment requires more than the current observation, as execution reliability depends on recent progress, prior outcomes, and the trajectory leading to the current state. We therefore employ a large language model $\pi_{\phi}$ as the critic, that, conditioned on the shared context $x_t$ and a local proposal $\hat a_t^L$, assesses whether local execution remains reliable enough to continue and assigns one of two control labels, \texttt{local} or \texttt{cloud}. Let
\begin{equation}
    s_t^m = \log P_{\phi}\left(y=m \mid x_t,\hat a_t^L\right),
    \qquad m\in\{L,C\},
\end{equation}
denote the log-likelihood assigned to each candidate label. Their relative likelihood is normalized into a reliability-aware confidence score
\begin{equation}
    c_t = \frac{\exp(s_t^L)}{\exp(s_t^L)+\exp(s_t^C)}.
    \label{eq:critic_score}
\end{equation}
A larger $c_t$ indicates stronger evidence for retaining or returning control to the local executor. Further details on the critic model and implementation setting can be found in \S\ref{sec:setup}.

\paragraph{Stateful Control Transfer.}
The critic score provides an instantaneous estimate of whether local execution should continue, while the control state determines how this signal affects the ongoing trajectory. We use two asymmetric thresholds to govern escalation and return: 
\begin{equation}
    z_t =
    \begin{cases}
    C, & z_{t-1}=L \wedge  c_t \leq \theta_{\mathrm{esc}}, \\
    L, & z_{t-1}=C \wedge  c_t \geq \theta_{\mathrm{ret}}, \\
    z_{t-1}, & \text{otherwise}.
    \end{cases}
    \label{eq:switch}
\end{equation}
The thresholds satisfy $\theta_{\mathrm{esc}} < \theta_{\mathrm{ret}}$, introducing hysteresis between escalation and return. Once escalated, the cloud executor retains control across subsequent actions until local reliability recovers above $\theta_{\mathrm{ret}}$, avoiding frequent handoffs during multi-step recovery. While cloud control is active, the critic continues to evaluate local proposals to determine when local execution can safely resume.

\paragraph{Speculative Execution.}
During local execution, cloud action generation may additionally be launched speculatively alongside critic evaluation. If local control is retained, the cloud proposal is discarded; if escalation is triggered, the proposal is directly reused, reducing the latency of control transfer without changing the critic score or switching policy. In practice, the critic is invoked every k steps, where k is a per-benchmark setting reported in \S\ref{sec:setup}; this adds only modest overhead across the P5–P95 latency percentiles (\S\ref{sec:ablation}).

\begin{algorithm}[t]
\caption{Speculative Critic-Guided Adaptive Execution}
\label{alg:adaptive_execution}
\begin{algorithmic}[1]
\Require Instruction $G$, plan $P$, initial observation $o_0$,
executors $\pi_L,\pi_C$, critic $\pi_\phi$,
parameters $\theta_{\mathrm{esc}},\theta_{\mathrm{ret}}$

\State $z \gets L$, $M \gets \emptyset$

\For{each interaction step $t$}
    \State $x_t \gets (G,P,o_t,M)$

    \If{$t$ is selected for critic evaluation}
        \Statex \hspace{\algorithmicindent}
        \textit{// Assess and transfer control}
        \State $r_t^C \gets \Call{Async}{\pi_C,x_t}$
        \Comment{speculative backup when $z=L$}
        \State $\hat a_t^L \gets \pi_L(x_t)$
        \State $c_t \gets
        \Call{CriticConfidence}{x_t,\hat a_t^L}$
        \State $z \gets \Call{UpdateState}{z, c_t}$
        \Comment{Eq.~\ref{eq:switch}}

        \Statex \hspace{\algorithmicindent}
        \textit{// Commit the proposal selected by the new state}
        \If{$z=C$}
            \State $a_t \gets \Call{Await}{r_t^C}$
        \Else
            \State $a_t \gets \hat a_t^L$
            \State \Call{Discard}{$r_t^C$}
        \EndIf
    \Else
        \State $a_t \gets \pi_z(x_t)$
        \Comment{retain the current executor}
    \EndIf

    \State $(o_{t+1},e_t) \gets \Call{Execute}{a_t}$
    \State $M \gets
    \Call{UpdateMemory}{M,(o_t,a_t,e_t,o_{t+1})}$
\EndFor
\end{algorithmic}
\end{algorithm}

\subsection{Shared Handoff Memory}
\label{sec:memory}

Control transfer creates a state-continuity problem that neither executor can solve alone. An incoming executor observes the current state, but not which subgoals are already complete, which actions have already failed, or how the current state was reached. Lacking this context, it may repeat resolved work or re-attempt a known failure — precisely the behavior escalation is meant to correct. We therefore externalize execution context into a shared handoff memory $M_t$ that is owned by the controller rather than by either executor, and is accessible to both executors and the critic.

\paragraph{Execution Context.}
The shared memory captures execution information at two temporal scales:
\begin{equation}
M_t =
\left(
M_t^{\mathrm{obs}},
M_t^{\mathrm{epi}}
\right),
\label{eq:memory}
\end{equation}
where $M_t^{\mathrm{obs}}$ maintains a bounded window of recent observation--action--outcome transitions, providing fine-grained context for how the current state was reached. In contrast, $M_t^{\mathrm{epi}}$ stores a compact episode-level summary of verified progress, persistent failures, and other execution-relevant information that should remain available beyond the recent window. The two are complementary rather than redundant: a bounded window alone would discard early progress over long horizons, while a summary alone would lose the local context needed to interpret a recent failure. This decomposition preserves both without retaining the full interaction history.

\paragraph{Memory Updates.} Updates require no additional model call, so memory contributes no inference latency or cost. After each executed action, the transition is normalized and appended to the window; only state supported by the resulting observation or public execution feedback is promoted to $M_t^{epi}$, where newer verified state supersedes incompatible older records and unresolved failures persist until an observation indicates recovery. Unexecuted proposals, model reasoning traces, and privileged simulator state are never stored: memory grants no information unavailable to a single-executor baseline, only continuity across a change of executor.

\paragraph{Handoff-Consistent Access.}
%After each environment interaction, $M_t^{\mathrm{obs}}$ is updated with the latest transition, while newly established progress or persistent failures are incorporated into $M_t^{\mathrm{epi}}$. Importantly, a control transfer does not reset or reconstruct the memory state. The same updated memory is exposed to $\pi_L$, $\pi_C$, and $\pi_\phi$, so switching the active executor changes only the control state $z_t$ while preserving the execution context used for subsequent actions and reliability assessment. This enables the incoming executor to resume from the existing task state rather than repeating previously attempted behavior or reconstructing prior progress. We empirically examine the effect of shared handoff memory in \S\ref{sec:ablation}.
At a control transfer, the same serialized memory is exposed to $\pi_L$, $\pi_C$, and $\pi_\phi$; it is neither reset nor regenerated for the incoming owner. A handoff therefore changes only the control state $z_t$, leaving the execution context intact, so the incoming executor resumes from the existing task state rather than reconstructing prior progress. This access is bidirectional. For example, if the local executor opens a cabinet but repeatedly fails to grasp the target, the cloud executor receives both the verified open state in $M_t^{\mathrm{epi}}$ and the recent failure sequence in $M_t^{\mathrm{obs}}$. It can therefore begin with recovery rather than repeat completed navigation and manipulation. Actions performed during cloud control update the same memory before control returns, allowing the local executor to continue from the post-recovery state. Section~\ref{sec:ablation} evaluates how each component affects task recovery and execution continuity.

\section{Evaluation}

\subsection{Experimental Setup}
\label{sec:setup}

\paragraph{Models and implementation.}
We use Claude Sonnet 4.6~\cite{ClaudeSonnet46} as the cloud planner and executor and a single zero-shot Qwen3.6-35B-A3B~\cite{Qwen36} checkpoint as both the local executor and the critic, served with SGLang on H200 GPUs and differing only in prompt and output interface. The critic evaluation interval is k = 1 on Habitat and k = 3 on EmbodiedBench and VLABench. Generation uses temperature 0 throughout, with a fixed seed for the local models. Escalation and return thresholds are selected per benchmark from a sweep over $(\theta_{\mathrm{esc}}, \theta_{\mathrm{ret}})$.
%, yielding $(0.18, 0.55)$, $(0.40, 0.55)$, and $(0.18, 0.30)$ on Habitat, EmbodiedBench, and VLABench; 
\S\ref{sec:understanding_escalation} reports the full sweep on Habitat and shows that performance is stable across a broad band of settings. All methods use the same models and decoding settings and differ only in how these models are coordinated during execution. \S\ref{sec:ablation} additionally evaluates heterogeneous executor--critic pairings with Qwen3-VL-8B~\cite{Qwen3VL} and InternVL3.5-14B~\cite{InternVL}.

\begin{table*}[t]
    \centering
    \caption{\capthead{Reliability--efficiency performance across long-horizon embodied benchmarks.} 
    %$^\ast$ denotes adaptations to our common benchmark interfaces. %Best results are shown in \textbf{bold}. 
    The italic row reports \name relative to Cloud-only (points for SR and CRR, percent for E2E and TBA).
    \name achieves the strongest reliability--latency trade-off among collaboration methods, and is the only one to exceed 70\% success on all three benchmarks.}
    \label{tab:main_results}
\setlength{\tabcolsep}{7.5pt}
\renewcommand{\arraystretch}{1.15}

\begin{tabular}{l cccc cccc cccc}
\toprule
& \multicolumn{4}{c}{\textbf{Habitat}}
& \multicolumn{4}{c}{\textbf{EmbodiedBench}}
& \multicolumn{4}{c}{\textbf{VLABench}} \\

\cmidrule(lr){2-5}
\cmidrule(lr){6-9}
\cmidrule(lr){10-13}

    \textbf{Method}
    & \textbf{SR} $\uparrow$
    & \textbf{CRR} $\downarrow$
    & \textbf{E2E} $\downarrow$
    & \textbf{TBA} $\downarrow$
    & \textbf{SR} $\uparrow$
    & \textbf{CRR} $\downarrow$
    & \textbf{E2E} $\downarrow$
    & \textbf{TBA} $\downarrow$
    & \textbf{SR} $\uparrow$
    & \textbf{CRR} $\downarrow$
    & \textbf{E2E} $\downarrow$
    & \textbf{TBA} $\downarrow$ \\

    & (\%) & (\%) & (s) & (s)
    & (\%) & (\%) & (s) & (s)
    & (\%) & (\%) & (s) & (s) \\

    \midrule

    \rowcolor{gray!13}
    \multicolumn{13}{l}{\textit{Single-executor baselines}} \\

    Local-only
    & 64.2 & -- & 10.56 & 0.78
    & 40.0 & -- & 11.25 & 0.85
    & 71.2 & -- & 16.36 & 0.53 \\

    Local-ReAct
    & 60.8 & -- & 22.99 & 1.58
    & 58.8 & -- & 17.32 & 1.19
    & 67.5 & -- & 17.28 & 0.85 \\

    Cloud-only
    & 91.7 & 100.0 & 49.62 & 4.61
    & \textbf{80.0} & 100.0 & 40.07 & 4.41
    & 85.0 & 100.0 & 23.67 & 2.47 \\

    Cloud-ReAct
    & 88.3 & 100.0 & 53.52 & 6.14
    & \textbf{80.0} & 100.0 & 77.47 & 6.88
    & 82.5 & 100.0 & 28.17 & 4.44 \\

    \addlinespace[2pt]

    \rowcolor{gray!13}
    \multicolumn{13}{l}{\textit{Collaboration baselines}} \\

    RouteLLM$^\ast$
    & 55.8 & 27.4 & 32.18 & 0.79
    & 53.8 & 27.0 & 33.03 & 1.14
    & 73.8 & 11.7 & 18.67 & 0.54 \\

    Hi Robot$^\ast$
    & 65.0 & 4.0 & 15.54 & 0.77
    & 52.5 & 9.0 & 18.22 & 0.94
    & 76.2 & 25.6 & 24.48 & 0.58 \\

    AdaPlanner
    & 74.2 & 2.2 & 21.66 & 1.26
    & 58.8 & 18.2 & 18.19 & 0.97
    & 81.2 & 37.3 & 25.42 & 3.90 \\

    \addlinespace[2pt]
    \midrule

    \rowcolor{blue!7}
    \textbf{\name}
    & \textbf{92.5} & 24.4 & 32.36 & 0.80
    & 75.0 & 34.4 & 37.28 & 3.14
    & \textbf{87.5} & 33.5 & 22.62 & 1.31 \\
    \addlinespace[1pt]
    \textit{\quad vs.\ Cloud-only}
    & \textit{+0.8} & \textit{--75.6} & \textit{--34.8\%} & \textit{--82.6\%}
    & \textit{--5.0} & \textit{--65.6} & \textit{--7.0\%} & \textit{--28.8\%}
    & \textit{+2.5} & \textit{--66.5} & \textit{--4.4\%} & \textit{--47.0\%} \\

    \bottomrule
\end{tabular}
\end{table*}

\paragraph{Baselines.}
We compare against two classes of methods. \emph{Single-executor baselines} include Local-only and Cloud-only, together with their ReAct-augmented variants, Local-ReAct and Cloud-ReAct~\cite{ReAct}. \emph{Collaboration baselines} cover complementary coordination strategies: Hi Robot*~\cite{HiRobot} represents a fixed cloud-planner--local-executor hierarchy, RouteLLM*~\cite{Routellm} represents model routing during execution, and AdaPlanner~\cite{AdaPlanner} represents feedback-triggered cloud replanning with local execution. Hi Robot* and RouteLLM* denote benchmark-specific adaptations of the original methods.
%implementation details are provided in Appendix~\ref{app-sec:baseline-adaptations}. 
``Local-only'' and ``Cloud-only'' refer to the executor rather than to the planning model. All baselines receive the same memory representation; for methods that transfer requests between executors (RouteLLM*), the memory is passed intact to the receiving model, identical to \name.

\paragraph{Benchmarks.}
We evaluate on 120 Habitat~\cite{Habitat} rearrangement instances, 80 instances from EmbodiedBench~\cite{Embodiedbench}, and 80 VLABench~\cite{VLABench} instances, covering mobile manipulation, long-horizon household interaction, and tabletop manipulation, respectively. The three benchmarks expose substantially different execution interfaces, including grounded PDDL-style actions, discrete action IDs, and structured macro skills, that allow us to test whether the control mechanism generalizes across different action abstractions and embodiment settings. All methods are evaluated on the same instances.

\paragraph{Metrics.}
We report task success rate (SR), median end-to-end episode latency (E2E), median time between actions (TBA), average trajectory length (Steps), and cloud request ratio (CRR). E2E measures wall-clock time from task issuance to episode termination, while TBA captures closed-loop interaction responsiveness. We define $\mathrm{CRR}=N_C/(N_C+N_L)$, where $N_C$ and $N_L$ count cloud and local \emph{executor} requests, respectively. All cloud executor requests, including speculative requests, are included in $N_C$. Critic calls are excluded from CRR and reported separately in the latency breakdown analysis (\S\ref{sec:ablation}).

\subsection{Overall Performance}
\label{sec:overall}

\paragraph{\name maintains task success while improving closed-loop responsiveness.}
Table~\ref{tab:main_results} reports task completion and closed-loop response times across the three benchmarks. Relative to Local-only, \name improves success rate by $16.3$--$35.0$ percentage points. It further exceeds Cloud-only success on Habitat and VLABench, while remaining within $5.0$ points on EmbodiedBench. The slight gains over Cloud-only can arise from differences in execution trajectories: \name may reach a different environment state before invoking the cloud executor, while continuous cloud execution may remain trapped in an ineffective action pattern. Meanwhile, \name reduces median TBA relative to Cloud-only by $28.8$--$82.6\%$ and lowers median E2E latency across all three benchmarks. These reductions allow the agent to act more responsively without substantially sacrificing task completion.

\begin{figure*}[t]
    \centering
    \includegraphics[width=0.98\textwidth]
    {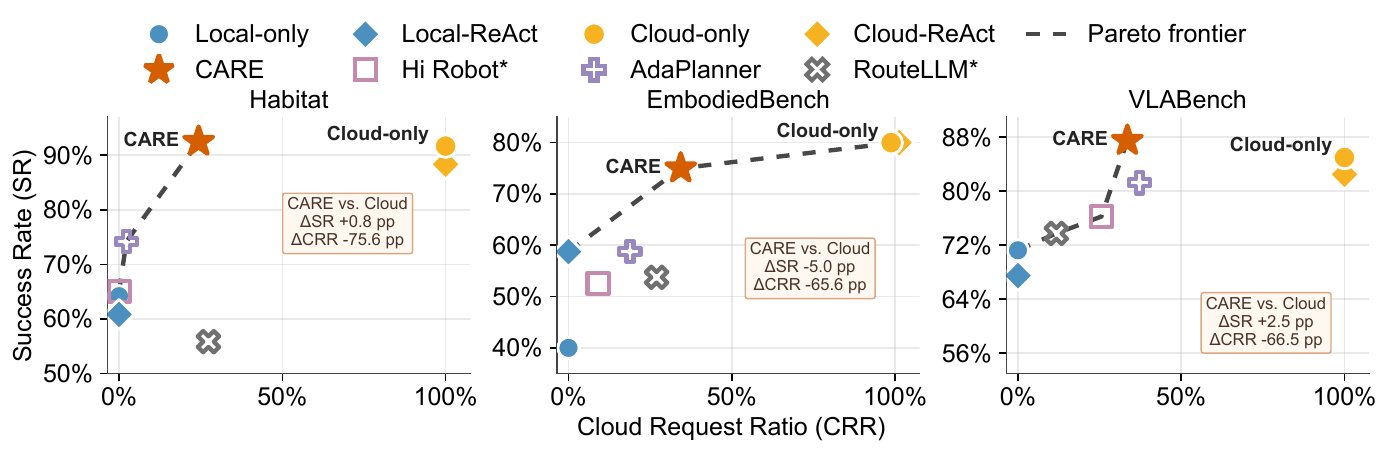}
    \vspace{-3pt}
    \caption{\capthead{Success rate versus cloud request ratio across three embodied benchmarks.} Cloud request ratio denotes the fraction of executor-model requests generated by the cloud model. Dashed lines indicate the empirical non-dominated frontier among the evaluated methods. \name approaches or exceeds Cloud-only success with substantially fewer cloud-executed decisions.}
    \label{fig:success_cloud_tradeoff}
\end{figure*}

\paragraph{\name achieves high task success with selective cloud use.}
Figure~\ref{fig:success_cloud_tradeoff} plots success against cloud request ratio, revealing a different aspect of the comparison. Cloud-only and Cloud-ReAct occupy the far-right edge at $100\%$ CRR, whereas \name uses the cloud for only $24.4$--$34.4\%$ of executor decisions and remains on the empirical non-dominated frontier in all three panels. Hi Robot, RouteLLM, and AdaPlanner use comparable or fewer cloud requests but consistently lie below \name in task success. This pattern suggests that more cloud requests alone do not guarantee higher task success; \name's advantage comes from using them at execution states where stronger reasoning is most useful, as examined in \S\ref{sec:understanding_escalation}.

\paragraph{\name shifts toward cloud assistance when local execution is less capable.}
EmbodiedBench illustrates how this balance changes with the local executor. Local-only achieves $40.0\%$ success, the lowest among the three benchmarks. Accordingly, \name uses a more aggressive escalation threshold and invokes the cloud more frequently. Its action latency therefore moves away from the local but remains below Cloud-only, while recovering substantially more task success than Local-only. Thus, \name does not enforce a fixed cloud-use target: its responsiveness depends on how much of the trajectory can be handled locally.

\begin{figure*}[t]
    \centering

    \begin{subfigure}[t]{0.38\linewidth}
        \centering
        \includegraphics[width=\linewidth]{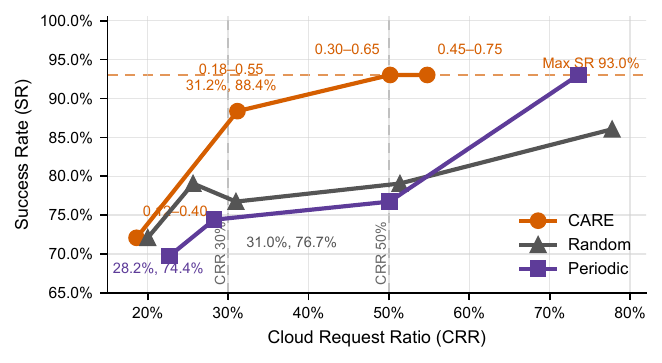}
        \caption{Matched-budget escalation.}
        \label{fig:matched_budget}
    \end{subfigure}\hfill
    \begin{subfigure}[t]{0.30\linewidth}
        \centering
        \includegraphics[width=\linewidth]{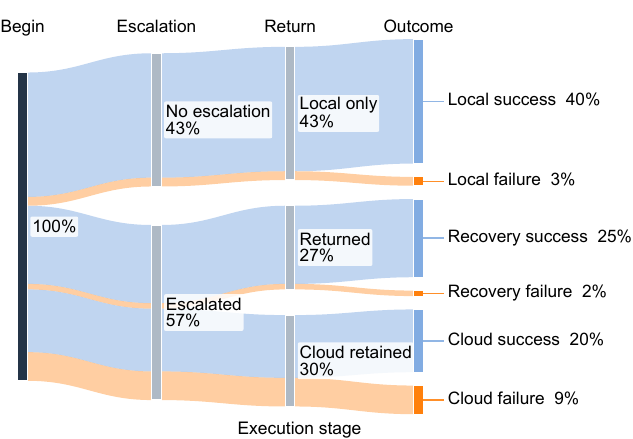}
        \caption{Episode flow.}
        \label{fig:episode_flow}
    \end{subfigure}\hfill
    \begin{subfigure}[t]{0.26\linewidth}
        \centering
        \includegraphics[width=\linewidth]{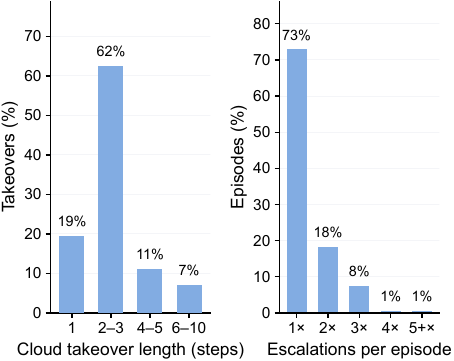}
        \caption{Takeover duration.}
        \label{fig:takeover_stats}
    \end{subfigure}

    \caption{\capthead{Understanding adaptive escalation.}
    (a) \name achieves higher success than matched-budget random and periodic escalation baselines.
    (b) Episode-level control pathways: 43\% of episodes never escalate and 93\% of those succeed; among escalated episodes, returned ones succeed at the same 93\% rate.
    (c) Most takeovers span multiple decisions, while most escalated episodes contain only one takeover.}
    \label{fig:adaptive_escalation}
\end{figure*}

\subsection{Does \name Know When to Escalate?}
\label{sec:understanding_escalation}

We evaluate the title claim from two perspectives: whether critic-guided escalation improves task success under matched cloud usage, and how control transfers unfold over complete episodes.

\paragraph{Timely escalation improves recovery from local failures.}
To isolate cases in which cloud intervention can change the outcome, we evaluate Habitat instances on which Local-only fails. We sweep $(\theta_{\mathrm{esc}},\theta_{\mathrm{ret}})$ to vary \name's cloud request ratio (CRR) and compare it with Random and Periodic schedules under matched cloud usage (Figure~\ref{fig:matched_budget}). Under a tight ${\sim}20\%$ CRR, all three methods achieve similar success, indicating that selective routing provides little additional benefit when the intervention budget is too restrictive. Beyond this regime, \name converts additional cloud budget into substantially stronger recovery: over the 30--50\% CRR range, its success rises from 88.4\% to 93.0\%, while the state-independent schedules remain between 74.4\% and 79.0\%. Increasing cloud usage alone does not close this gap: Periodic requires over 70\% CRR to reach 93.0\%, while Random remains below it even near 80\% CRR. \name itself saturates at 93.0\% around 50\% CRR, showing diminishing returns from more aggressive thresholds. Thus, successful recovery requires sufficient cloud capacity, but \name's advantage comes from allocating that capacity to the execution states where intervention is most useful.

\paragraph{\name retains local control when possible and sustains cloud control when needed.}
Figure~\ref{fig:episode_flow} shows that 43\% of episodes complete without escalation, of which 93\% succeed, indicating that \name avoids unnecessary cloud intervention when local execution is sufficient. Episodes returned to local control achieve the same 93\% success rate, suggesting that local execution is resumed without sacrificing reliability. Meanwhile, 30\% of episodes remain under cloud control until termination, and Figure~\ref{fig:takeover_stats} shows that 81\% of takeovers span multiple decisions. Together, these results show that \name produces state-dependent control transfers rather than independent per-step model choices.

%\paragraph{Critic confidence predicts local execution outcomes.}
%We further examine whether the critic score is informative of local executability. Across 325 retained-local checkpoints, $c_t$ achieves an AUROC of 0.717 (95\% CI: 0.638--0.793) for distinguishing successful from unsuccessful local continuation. Local execution succeeds at 98.0\% of checkpoints with $c_t \geq 0.55$, compared with 87.5\% when $0.18 \leq c_t < 0.55$. Although the critic is not a perfect predictor, its confidence provides a meaningful signal for deciding when local execution should continue.

\subsection{System Analysis and Ablation Studies}
\label{sec:ablation}

\begin{figure}
    \centering
    \includegraphics[width=\linewidth]{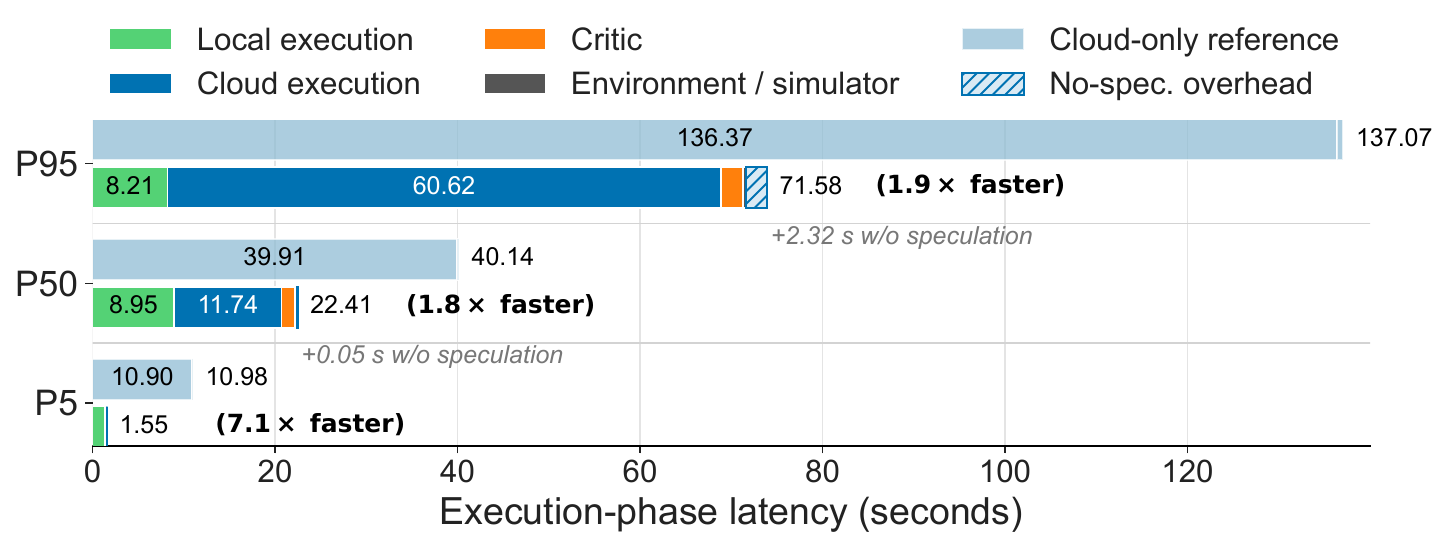}
    \caption{\capthead{Task execution latency breakdown on Habitat.} Across the P5, P50, and P95 episodes, \name reduces latency by $1.8$--$7.1\times$ relative to Cloud-only, while critic evaluation and speculative execution add only modest overhead.}
    \label{fig:latency_breakdown}
\end{figure}

\paragraph{\name reduces task execution time across the latency distribution.}
All methods issue the same one-time planning call, so Figure~\ref{fig:latency_breakdown} reports execution-phase latency with this shared cost excluded. \name reduces execution latency across the P5–P95 range relative to Cloud-only, indicating that its average-case gains do not come at the expense of tail latency. The reduction comes from replacing cloud execution with lower-latency local execution rather than from shortening trajectories. Critic evaluation contributes only 0.15–2.45 s across the measured percentiles, and speculative control transfer overlaps cloud generation with critic evaluation so that escalation avoids a serial cloud round trip: removing speculation costs 0.05 s at P50 and 2.32 s at P95. \name therefore improves the latency distribution without introducing substantial control overhead.

\paragraph{Shared handoff memory preserves task progress across control transfers.}
As shown in Table~\ref{tab:memory_ablation}, we resume execution from recorded control-transfer points in Habitat, sampled from states where the critic triggered escalation and therefore representative of the difficult states at which handoffs actually occur. Holding the environment state and execution history fixed, we vary only the memory available to the incoming executor. Removing memory reduces success from 100\% to 25\% and more than doubles both trajectory length and cloud usage, indicating a substantial loss of progress continuity. Episodic memory alone recovers most of the success (90\%), while observation memory alone recovers less (75\%); combining them restores full success and further reduces execution cost. The two components therefore degrade in the distinguishable ways predicted in \S\ref{sec:memory}: verified task progress is what an incoming executor most needs, and recent action–outcome context supplies the complementary local detail.

\paragraph{\name remains effective across executor and critic choices.}
To assess whether \name relies on a particular executor--critic pairing, we replace one component at a time while keeping the other fixed. Table~\ref{tab:model_robustness} shows that the default pairing achieves the best success rate, trajectory length, and end-to-end latency. Nevertheless, all substituted configurations retain 83.3--88.3\% success, indicating that \name remains effective across different model families rather than depending on a single pairing. Executor substitutions incur larger increases in trajectory length and latency, whereas critic substitutions preserve execution efficiency more closely. In particular, the 8B critic retains 86.7\% success with substantially lower degradation than the executor replacements, suggesting that critic capacity can be reduced when deployment cost is prioritized. 
%Overall, \name benefits from the strongest default pairing while still supporting flexible executor--critic choices.

\begin{table}[t]
    \centering
    \small
    \setlength{\tabcolsep}{3.5pt}
    \renewcommand{\arraystretch}{1.08}
    \caption{\capthead{Effect of handoff memory at matched Habitat execution states ($N=20$).} Variants differ only in the memory available to the incoming executor.}
    \label{tab:memory_ablation}
    \begin{tabular}{lcccc}
        \toprule
        \textbf{Incoming Memory}
        & \textbf{SR (\%) $\uparrow$}
        & \textbf{Steps $\downarrow$}
        & \textbf{E2E (s) $\downarrow$}
        & \textbf{Cloud $\downarrow$} \\
        \midrule

        Full
        & \textbf{100.0}
        & \textbf{12.15}
        & \textbf{38.83}
        & \textbf{109} \\

        Episodic only
        & 90.0
        & 16.05
        & 61.80
        & 152 \\

        Observation only
        & 75.0
        & 15.80
        & 63.18
        & 144 \\

        None
        & 25.0
        & 25.15
        & 89.52
        & 245 \\

        \bottomrule
    \end{tabular}
\end{table}

\begin{table}[t]
    \centering
    \small
    \setlength{\tabcolsep}{3.5pt}
    \renewcommand{\arraystretch}{1.08}
    \caption{\capthead{Executor and critic substitutions on Habitat.} The default uses Qwen3.6-35B-A3B for both components; each remaining row replaces one component.}
    \label{tab:model_robustness}
    \begin{tabular}{llccc}
        \toprule
        \textbf{Variant}
        & \textbf{Model}
        & \textbf{SR (\%) $\uparrow$}
        & \textbf{Steps $\downarrow$}
        & \textbf{E2E (s) $\downarrow$} \\
        \midrule

        \rowcolor{gray!10}
        \textbf{Default}
        & Qwen3.6-35B-A3B
        & \textbf{92.5}
        & \textbf{10.43}
        & \textbf{32.36} \\

        Executor
        & Qwen3-VL-8B
        & 85.0
        & 13.32
        & 51.18 \\

        Executor
        & InternVL3.5-14B
        & 88.3
        & 15.20
        & 64.83 \\

        Critic
        & Qwen3-VL-8B
        & 86.7
        & 11.43
        & 40.55 \\

        Critic
        & InternVL3.5-14B
        & 83.3
        & 11.57
        & 45.25 \\

        \bottomrule
    \end{tabular}
\end{table}

\section{Limitations and Future Work}
\name operates at the embodied decision layer above the underlying low-level controller; however, its behavior under sensing noise, network variability, and hardware execution failures remains unverified. Accordingly, the reported latencies characterize decision-layer responsiveness rather than end-to-end physical robot response. Evaluating \name on a physical platform is therefore an important next step, particularly because real network conditions may affect the latency and availability of cloud assistance. Beyond physical deployment, systematically characterizing how model scale, task difficulty, and action abstraction shape the boundary between local and cloud competence remains an important future direction.
%The critic assesses local continuation from a single proposed action and shared execution history, without explicit multi-step lookahead, and may therefore miss failures that become apparent only after several actions. 
%Finally, escalation thresholds are fixed per benchmark; adapting them online as executor capability and operating conditions change remains future work.

\section{Conclusion}
We presented \name, an adaptive execution framework for cloud--local embodied agents. \name evaluates local proposals against the current observation and execution history, transfers control to a stronger cloud executor when local execution deteriorates, and preserves task state across model handoffs. Across Habitat, EmbodiedBench, and VLABench, \name improves task success by $6.3$--$18.3$ percentage points over the strongest collaboration baselines while reducing median time-between-action latency by $28.8$--$82.6\%$ relative to Cloud-only. Matched-budget comparisons further show that these gains arise from selecting useful intervention states rather than simply increasing cloud usage. Overall, executor selection should adapt with the embodied trajectory rather than remain fixed throughout a task.
%Future work will extend \name to larger executor pools, explicit latency and cost budgets, and real-world robotic systems.

%\subsubsection*{Author Contributions}
%If you'd like to, you may include  a section for author contributions as is done
%in many journals. This is optional and at the discretion of the authors.

%\subsubsection*{Acknowledgments}
%Use unnumbered third level headings for the acknowledgments. All
%acknowledgments, including those to funding agencies, go at the end of the paper.
\bibliography{main}
\bibliographystyle{IEEEtran}

%\newpage
%\appendix

%\input{Sections/Appendix}

\end{document}